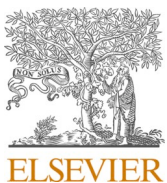
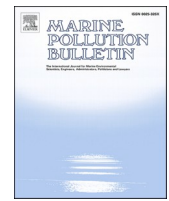

# Riverbed litter monitoring using consumer-grade aerial-aquatic speedy scanner (AASS) and deep learning based super-resolution reconstruction and detection network

Fan Zhao [a,*], Yongying Liu [a], Jiaqi Wang [a], Yijia Chen [a], Dianhan Xi [a], Xinlei Shao [b], Shigeru Tabeta [a], Katsunori Mizuno [a,*]

[a] *Department of Environment Systems, Graduate School of Frontier Sciences, The University of Tokyo, Japan*
[b] *Department of Socio-Cultural Environmental Studies, Graduate School of Frontier Sciences, The University of Tokyo, Japan*

ABSTRACT

Underwater litter is widely spread across aquatic environments such as lakes, rivers, and oceans, significantly impacting natural ecosystems. Current automated monitoring technologies for detecting this litter face limitations in survey efficiency, cost, and environmental conditions, highlighting the need for efficient, consumer-grade technologies for automatic detection. This research introduces the Aerial-Aquatic Speedy Scanner (AASS) combined with Super-Resolution Reconstruction (SRR) and an enhanced YOLOv8 detection network. The AASS system boosts data acquisition efficiency over traditional methods, capturing high-resolution images that accurately identify and categorize underwater waste. The SRR technique enhances image quality by mitigating common issues like motion blur and low resolution, thereby improving the YOLOv8 model's detection capabilities. Specifically, the RCAN model achieved the highest mean average precision (mAP) of 78.6 % for object detection accuracy on reconstructed underwater litter among the tested SR models. With a magnification factor of 4, the SR test set shows an improved mAP compared to the Bicubic test set. These results demonstrate the effectiveness of the proposed method in detecting underwater litter.

## 1. Introduction

Unmanned Aerial Vehicles (UAVs) are extensively used for extracting and processing information in aquatic environments, such as lakes, rivers, and oceans, especially for monitoring aquatic life and waste (Geraeds et al., 2019). Data on waste density and distribution are crucial for effective waste management and play a significant role in aquatic life conservation and water quality management (Kowsari et al., 2023). Due to their high accuracy, efficiency, and cost-effectiveness, UAVs and Computer Vision (CV) technologies are widely adopted for waste detection (González-Sabbagh and Robles-Kelly, 2023; Sharma et al., 2023; Jakovljevic et al., 2020). However, these techniques primarily focus on surface waste, often overlooking substantial submerged waste. The quality of traditional remote sensing data, including sonar (Chai et al., 2023), UAVs (Liu et al., 2023; Merrifield et al., 2023), and satellite imagery (Bagwari et al., 2023; Kikaki et al., 2024; Palombi and Raimondi, 2022), is limited by water quality and depth. Efficiently collecting and processing high-quality images for realistic underwater waste detection remains a key challenge in aquatic waste identification.

Advancements in deep learning have enabled Convolutional Neural Networks (CNNs) to autonomously recognize image features in supervised learning scenarios without relying on the knowledge of domain experts. This progress means that CNN-processed images show enhanced accuracy and robustness in capturing texture features compared to traditional image processing methods (Simonyan and Zisserman, 2014). However, models trained this way struggle to identify targets in real, diverse underwater waste images, posing a challenge for CNNs to classify various types of underwater waste differing in shape, size, and color (Wu et al., 2023). While object detection algorithms can effectively recognize laboratory or simulated underwater waste, they fall short in providing accurate information about the real texture and categories of waste (Agustsson and Timofte, 2017; Majchrowska et al., 2022).

While CNN-based networks have achieved some success in underwater waste recognition, technical challenges such as image blur due to turbidity and deformation of waste, limit their field adoption. Environmental factors during image capture, such as the riverbed or seafloor background, lighting, and turbidity, can blur the edges and textures of underwater waste, directly affecting recognition and classification accuracy (Zhou et al., 2022). The image perspective also impacts recognition performance, as the front and top views of the same object can appear drastically different (Xiao et al., 2022). Additionally, waste can deform, become dirty, get wet, or overlap, altering its edges and textures. Thus, there is an increasing need for robust models and extensive training data (Duan et al., 2020). Remotely Operated Vehicles (ROVs) can collect large amounts of high-quality underwater waste images, but their high cost and low maneuverability make them unsuitable for large-scale underwater waste surveys and insufficient for the data requirements of computer vision-based underwater waste recognition models. Other towed underwater camera arrays system, like the Speedy Sea Scanner (SSS) (Mizuno et al., 2019) and Portable Speedy Sea Scanner (P-SSS) (Terayama et al., 2022) enhances the ability to capture large-scale underwater images. However, they require a towing ship and 2–3 professional operators, resulting in high costs and limited use in narrow inland rivers. Additionally, deploying and retrieving these systems demand significant investments in operator and vessel costs.

While UAVs are effective in conducting field surveys for underwater waste, significant image clarity challenges remain due to factors such as water clarity, surface reflection, attenuation, wind, and waves, impacting the application of deep learning models (Majchrowska et al., 2022). A review of recent advanced amphibious UAVs in recent years, such as Aerial-Aquatic Robots (Li et al., 2022), Dipper (Rockenbauer et al., 2021), and HAMADORI 6000 (Space Entertainment Laboratory Co., Ltd, n.d.), reveals that these devices are either prohibitively expensive or still in experimental stages, making them impractical for real-world use. This study builds upon these limitations by focusing on low-cost, consumer-grade technology to improve accessibility.

The accuracy of automatic underwater litter detection using UAV images is highly dependent on image resolution. However, almost all UAVs are affected by the survey conditions and have insufficient resolution, which seriously affects the accuracy of object detection (Zhao et al., 2023a, b). Therefore, enhancing the resolution of image resolution is crucial for improving detection accuracy. Super-Resolution Reconstruction (SRR) is a widely used and effective method in underwater target detection (Anwar and Li, 2020; Jin et al., 2021). Numerous traditional image processing methods, such as unsharp mask filtering, median filtering, and histogram equalization (Kasim et al., 2021), are available for enhancing image clarity. Yet, these methods primarily focus on improving image quality by utilizing existing pixels, offering limited resolution enhancement, which is not significantly beneficial for the efficiency of waste classification or detection models (Raveendran et al., 2021). SRR, an emerging technology, addresses issues like motion blur and low resolution. It includes three algorithmic categories: interpolation, reconstruction, and machine learning (Li et al., 2021). Each SRR method has drawbacks. Interpolation-based methods focusing merely on pixel point manipulation, often results in blurred images from excessive detail reduction. Reconstruction-based approaches, while integrating prior knowledge, fall short in reconstructing texture-rich images (Wang et al., 2015). Traditional machine learning algorithms provide more precise outcomes but are time-intensive and challenging to optimize (Zhao et al., 2018). Deep learning-based SRR has shown promise in overcoming underwater object detection and classification challenges (Wang et al., 2020). The first deep learning network for SRR was the Super-Resolution Convolutional Neural Network (SRCNN) (Chao et al., 2014), followed by advanced networks like Super-Resolution Generative Adversarial Network (GAN)- based networks (Ledig et al., 2017). Research on applying deep learning-based SRR to underwater waste image processing is limited (Wang et al., 2020; Heidemann et al., 2012). Li et al. (2019a, b) showed that GAN models improved clarity in underwater fish images, which is further explore in this study. Another study demonstrated that an SRR combined with scattering and fusion methods could enhance underwater image quality, proving SRR's effectiveness in improving the quality of underwater images (Anwar and Li, 2020). These studies reveal SRR's potential to improve underwater image clarity, but research on its applications in underwater advanced tasks like object detection and semantic segmentation is sparse. Additionally, the impact of various SRR networks and their magnification factors on image reconstruction and detection accuracy in underwater environments remains unexplored. This study builds upon these findings to enhance the detection and classification of underwater litter using SRR techniques.

To tackle the challenges of underwater waste detection, in terms of hardware, this study introduces an innovative consumer-grade water surface rapid scanning remote sensing image acquisition system, i.e. the Aerial-Aquatic Speedy Scanner (AASS). The AASS integrates the efficiency and convenience of UAVs with the high-resolution imaging capabilities of ROVs. This hybrid system significantly reduces survey costs while enhancing data accuracy. Additionally, the AASS operates without the need for a support vessel, enabling real-time communication and GPS-based offshore operations. Optimized for depths ranging from 0.5 to 10 m, the AASS excels in shallow waters often rich in underwater litter and benthic organisms like coral reefs, which may struggle with maneuverability and detailed imaging in such environments. Traditional underwater vehicles face maneuverability issues in narrow rivers, highlighting the need for the AASS's flexibility. Using the AASS, this study efficiently collected data on underwater riverbed litter, facilitating automated surveys through object detection models. On the software side, this study proposes a novel approach combining deep learning-based SRR with an improved YOLOv8 model, applied to the collected data. Based on the collected field video data, panoramic riverbed waste maps were created using the Scale-Invariant Feature Transform (SIFT) method, a robust algorithm used for detecting and describing local features in images by identifying key points and computing their descriptors, which are invariant to image scaling, rotation, and partially invariant to changes in illumination and 3D camera viewpoint (Lowe, 2004). To train the super-resolution algorithms, original High-Resolution (HR) and Low-Resolution (LR) images were derived through slide-window cropping and down-sampling. The SR images generated from the super- resolution models enable detailed underwater waste analysis. The proposed YOLOv8, named RBL-YOLO, was trained using HR waste datasets and tested on both original and reconstructed SR image sets. Through comparative analysis and quantitative evaluation, this study assessed the models' effectiveness and the impact of different SRR magnification factors on image reconstruction and underwater litter detection.

This research aims to develop a cost-effective system for detecting underwater riverbed litter. It combines the AASS with deep learning-based SRR to improve image quality and detection accuracy. The AASS merges UAV agility with AUV high-resolution imaging for precise, affordable data collection. The study also introduces an enhanced RBL-YOLO detection network to more accurately identify and classify underwater litter, outperforming current methods.

## 2. Methodology

The methodological framework of this study is illustrated in Fig. 1. HR images were then produced by cropping the panorama into 512 × 512 pixel patches, which were labeled and enhanced for further analysis. A comparative analysis of object detection accuracy was then conducted across different SR image datasets to evaluate the performance of the same detection model on images produced by different SRR models. For an in-depth exploration of the methodologies employed, please refer to Sections 2.1–2.5.

### 2.1. Study site

The field data for this study was collected from the Hongqi River, a tributary of the Wujin River, located in Tonghai County within Yuxi City in Yunnan Province, China (see Fig. 2). This watershed, spanning approximately 40 km and covering around 200 km$^2$, is characterized by diverse hydrological and ecological features. It includes farmlands and highways and serves as a crucial resource for irrigation, domestic water supplies, and ecological conservation in Tonghai County. The Hongqi River significantly contributes to Lake Qilu's hydrological dynamics, accounting for 47.8 % of the lake's total water volume (Xing et al., 2018).

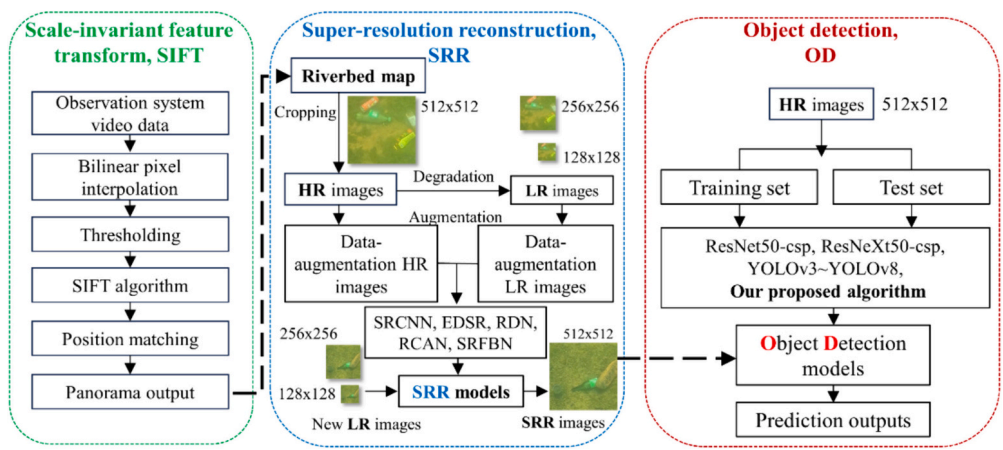

**Fig. 1.** Methodological framework of riverbed waste detection.

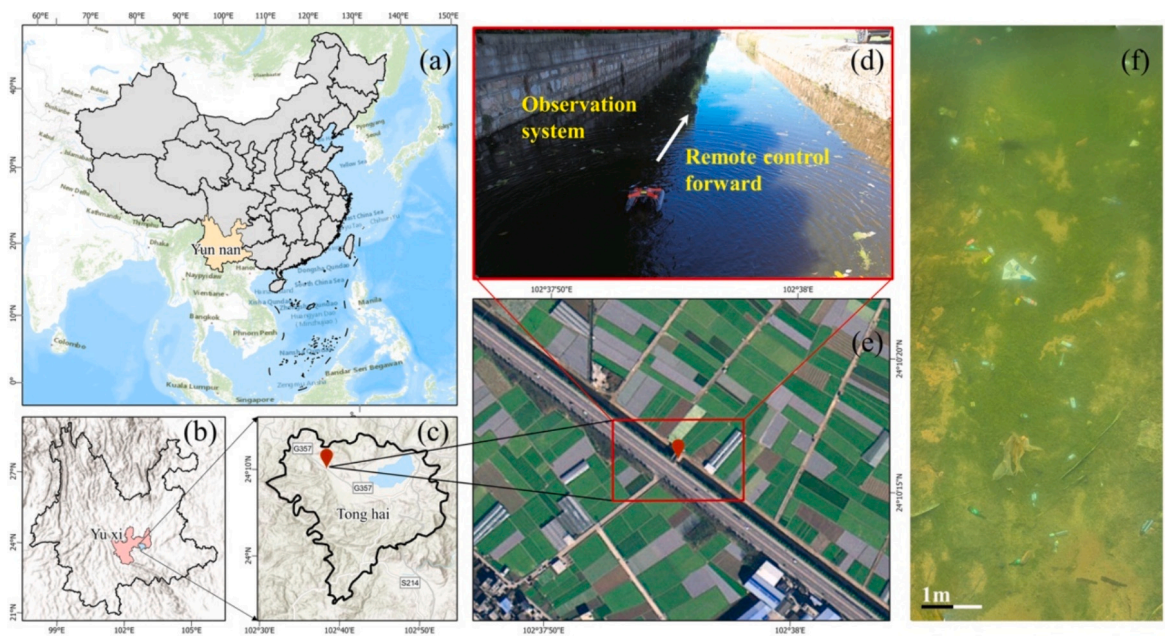

**Fig. 2.** Spatial representation of the research area: (a) Yunnan Province within China; (b-c) Tonghai County within Yuxi City and its location within Yunan Province; (d) Observation system scanning the riverbed; (e) Study region within the Hongqi River; (f) Sample of riverbed mapping generated using the SIFT algorithm.

*2.2. Data acquisition*

The observational device is equipped with an array of 4 K resolution underwater cameras mounted to a horizontal frame. The AASS, capable of both aerial and boat modes, was tested in a controlled pool environment (Fig. 3) to evaluate its performance metrics before field data collection. In boat mode, the camera-equipped frame submerges to a depth of approximately 40 cm below the water surface. The system commences operation at velocities ranging from 1 to 2 knots, making it ideal for detailed underwater surveys. This setup facilitates efficient underwater litter surveys, producing high-fidelity images of underwater environments. The AASS can be equipped with camera array beams of 2 m and 4 m lengths, allowing for versatile investigation of varying water conditions such as turbidity, watercourse width, target distribution characteristics, and specific survey objectives. The appropriate version of AASS can be flexibly selected to adapt to various survey areas. The enhanced clarity of these underwater images aids in the detection and categorization of underwater litter, providing a significant advantage over above-water monitoring methods. Additionally, employing its aerial mode, the system is capable of executing surveys of aquatic targets from an aerial perspective, incorporating the same capabilities as traditional UAVs. AASS integrates two survey modes and can be flexibly adjusted according to specific needs. Boat mode is ideal for surveys of benthos at individual level and morphology, and waters with poor local water quality, while flight mode is suitable for monitoring dynamic density changes, distribution, underwater litter traceability, and areas where water quality meets required standards. The flexible switching between modes ensures efficient deployment and versatile aquatic scientific surveys.

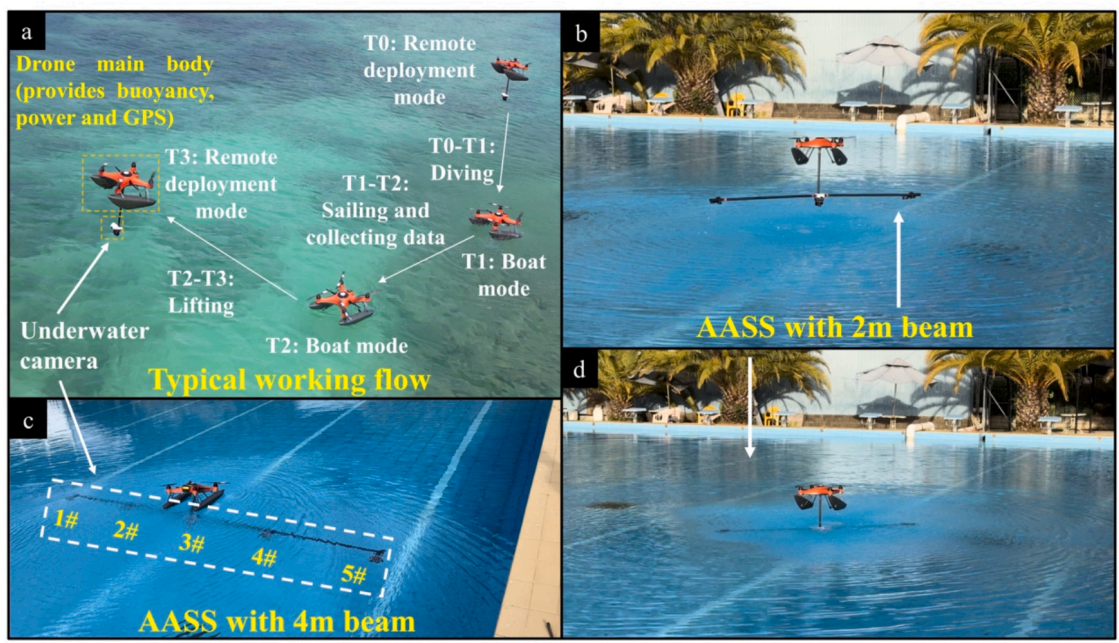

**Fig. 3.** AASS test in the marine and pool environments: (**a**) Typical workflow of single camera equipped with AASS in the sea trials; (**b**) and (**d**) show AASS switching between aerial and boat modes with the 2-meter extended beam and 3 underwater cameras; (**c**) depict AASS in boat mode with a 4-meter extended beam and 5 cameras.

*2.3. Data preprocessing and riverbed map generation*

To create a detailed depiction of the riverbed, the SIFT algorithm was utilized to integrate individual video frames captured during the survey (Lowe, 2004). Using footage from six of the eight video streams captured in the boat mode of the AASS, a detailed map of the riverbed was synthesized. Fig. 1 shows the detailed map generated using the SIFT algorithm.

*2.4. Super-resolution reconstruction based on deep learning*

SRR based on deep learning represents a cutting-edge approach in the field of image processing, aiming to enhance the resolution of low-quality images by generating high-quality counterparts. These pairs are generated by artificially downscaling HR images with a known image degradation model to simulate LR counterparts (Yang et al., 2022). The model is then trained to predict SR images from LR inputs by minimizing a loss function, which measures the difference between the predicted SR images and the original HR images in the training set. Selection of network architecture, hyperparameters, and optimization strategies is essential to SRR (Wang et al., 2020). The performance of the SRR model is evaluated based on how closely these reconstructed images match the quality and details of original HR images, using various evaluation metrics tailored to assess the fidelity and perceptual quality of the SR images (Zhang and Patel, 2018).

*2.4.1. Model structure*

In the field of SRR, numerous models have emerged, each introducing advancements through distinct network design approaches. Researchers have proposed models ranging from linear networks to generative adversarial networks, each addressing specific challenges in image reconstruction. Linear networks such as Super-Resolution Convolutional Neural Network (SRCNN) (Dong et al., 2014) offer simple structures. These simple networks face limitations in leveraging the breadth of image features for reconstructing new images. Evolving from linear designs, residual learning networks, including notable models like Enhanced Deep Super-Resolution Network (EDSR) (Lim et al., 2017), Residual Dense Network (RDN) (Yang et al., 2021), and Enhanced Residual Network (ERN), introduce a significant innovation to resolve the existing gradient degradation problem. These networks enable enhanced feature propagation, ensuring rapid convergence and strong learning capabilities. Further developments in this field have refined the architecture using dense connection networks, exemplified by the Super-Resolution Dense Network (SRDenseNet) and Deep Back-Projection Network (DBPN) (Haris et al., 2018). This dense connection strategy circumvents the vanishing gradient issue and allows for a reduction in model size without sacrificing performance. Recursive learning networks, including the Super-Resolution Feedback Network (SRFBN), integrate recursive learning strategies to iteratively refine image features. This recursive approach marks a significant stride in model improvement, offering a method for enhancing the accuracy of object detection on reconstructed images. The development of Generative Adversarial Networks (GANs), such as the Super-Resolution Generative Adversarial Network (SRGAN) (Ledig et al., 2017), Enhanced Super-Resolution Generative Adversarial Network (ESRGAN) (Wang et al., 2018), represents an advanced development in SRR. By utilizing adversarial training mechanisms, these models can produce high-quality images that set new benchmarks for realism in super-resolution tasks.

The SRCNN introduced a three-layer deep learning model that directly learns an end-to-end mapping between low and high-resolution images (Dong et al., 2014). This straightforward yet effective architecture significantly improved upon traditional interpolation methods. EDSR refined the approach by eliminating the batch normalization layer, thereby reducing memory usage by approximately 40 % during training (Lim et al., 2017). This allowed for the addition of more network layers and feature extraction capabilities, enhancing performance without increasing computational demand. RDN introduced the residual dense block, combining residual learning and dense connections for superior feature fusion and reuse, which provided a richer set of information for image reconstruction, allowing clarity and detail in output images (Zhang et al., 2018a, b). RCAN implemented channel attention mechanisms within its structure, focusing on the most relevant features for super-resolution. This attention mechanism improved the quality of the reconstructed images beyond what previous models had achieved (Zhang et al., 2018a, b). SRFBN employed a feedback mechanism through feedback blocks, enabling iterative refinement of image details. This recursive learning strategy allowed for continuous improvement of image quality, using high-level information to refine low-level features effectively (Li et al., 2019a, b). The ESRGAN evolved from SRGAN by integrating the residual in the residual dense block and adopting the relativistic average GAN. This two-step modification significantly enhanced the realism of SR images (Wang et al., 2018).

*2.4.2. Training parameters*

In the training phase, the model undergoes a process of parameter adjustment predicated on the computed discrepancy between each LR input image and its HR counterpart. For CNN-based architectures, parameter optimization is conducted utilizing the L1 loss function, which is subsequently minimized through the application of the Adam optimization algorithm (Johnson et al., 2016). For GAN-based models, optimization leverages both perceptual and adversarial losses, enhancing the authenticity and quality of the SR images (Wang et al., 2018).

In the conducted experiments, several key training parameters were established: the maximum number of training epochs was capped at 300, with a batch size of 16 and an initial learning rate set at 0.0001. To facilitate effective learning, a decay strategy for the learning rate was implemented, halving the rate every 200 epochs to gradually refine the learning process and ensure optimal convergence of the network parameters.

*2.4.3. Evaluation metrics*

The evaluation of SRR networks involves the utilization of evaluation metrics to assess the quality of reconstructed SR images compared to their HR counterparts. The Structural Similarity Index (SSIM) and Peak Signal-to-Noise Ratio (PSNR) are used as evaluation metrics in this study, which are widely recognized for their ability to gauge the perceptual and fidelity aspects of super-resolution images (Hore and Ziou, 2010). SSIM is designed to measure the visual impact of three characteristics of an image: luminance, contrast, and structure, which are critical for human visual perception. The SSIM index is calculated using the formula (1) (Wang et al., 2004).

$$SSIM(O,R) = \frac{(2\mu_O\mu_R + C_1)(2\sigma_{OR} + C_2)}{(\mu_O^2 + \mu_R^2 + C_1)(\sigma_O^2 + \sigma_R^2 + C_2)} \quad (1)$$

where $O$ and $R$ are the HR and SR image respectively, $\mu_O$ and $\mu_R$ denote their average luminance, $\sigma_O^2$ and $\sigma_R^2$ represents their variance respectively, $\sigma_{OR}$ represents the covariance, $C_1$ and $C_2$ are constants to stabilize division with weak denominator.

PSNR, on the other hand, is a measure derived from the mean squared error (MSE) between the HR and SR images. It reflects the ratio of the maximum possible power of a signal to the power of corrupting noise. The equation for PSNR is given by eq. (2) (Wang et al., 2003).

$$PSNR = 10 \times lg\left(\frac{MAX_I^2}{MSE}\right) \quad (2)$$

where $MAX_I$ represents the maximum pixel intensity of the image, and MSE is the mean squared error between the HR and SR images. Together, SSIM and PSNR provide comprehensive insights into the quality of SRR networks, encompassing both perceptual quality and fidelity, making them indispensable tools in the evaluation of super-resolution techniques.

*2.5. Object detection*

*2.5.1. Network architecture of RBL-YOLO*

The YOLO (You Only Look Once) series has consistently set benchmarks for speed and accuracy in object detection. YOLOv8 extends this tradition, offering substantial advancements in detection capabilities and computational efficiency. Its inference speed is also improved, achieving 44.9 milliseconds, making it suitable for real-time applications (Bochkovskiy et al., 2020; Redmon and Farhadi, 2018). Notwithstanding, studies indicate that YOLOv8, while effective in general object detection, struggles with the accurate detection of small objects due to inherent limitations in feature representation and spatial resolution, highlighting areas for further optimization (Shen et al., 2023).

This study proposes an enhancement strategy for YOLOv8, namely RBL-YOLO, by integrating three modifications: the RepVGG backbone (Ding et al., 2021), the SimAM attention mechanism (Yang et al., 2021), and the Efficient Intersection over Union (EIoU) loss function (Zheng et al., 2020). These modifications are specifically designed to augment the model's feature extraction capabilities and bounding box regression accuracy, with a particular focus on improving the detection of objects of various.

The integration of the RepVGG architecture as a backbone aims to streamline the convolutional network structure while either maintaining or enhancing its representational capacity (Dong et al., 2016). By utilizing a sequence of 3 × 3 convolutions and ReLU activations, this architectural substitution enhances the model's ability to effectively handle underwater debris datasets. These datasets are often characterized by uneven terrain, mixed with rocks, aquatic plants, and varying water quality, which are common features in riverbed environments. Moreover, RepVGG precisely captures subtle feature variations (Ding et al., 2021), facilitating the model's ability to learn and categorize various types of underwater debris accurately.

Moreover, the SimAM module is a lightweight self-attention mechanism, which can adaptively calibrate feature maps to prioritize spatial locations of significance within an image (Yang et al., 2021). This advantage aligns with the characteristics of underwater litter, which is rich in textures, edges, and colors. By emphasizing critical regions, the SimAM enables RBL-YOLO to better obtain contextual information, thus improving detection outcomes for objects that often blend into larger background features (Yang et al., 2021). The SimAM module enhances the stability of models on diverse underwater litter datasets by emphasizing crucial feature areas. It directs the model's focus toward regions of an image with substantial discriminative features, minimizing the impact of background noise. This attribute is particularly valuable when dealing with images that have complex backgrounds or multiple visual disturbances, as it aids in improving the model's ability to recognize relevant targets. As a lightweight attention mechanism, SimAM does not significantly increase computational demands (Li and Kang, 2024). This allows YOLOv8 to maintain operational speed while better capturing essential information, making it suitable for underwater litter detection tasks, which lays a foundation for real-time monitoring.

The replacement from complete Intersection over UnionCIoU (CIoU) to EIoU introduces an advanced metric for bounding box regression that incorporates both the traditional IoU score and the aspect ratio of predicted boxes relative to actual ground truth dimensions (Thulasya Naik et al., 2024). This enhancement not only ensures more precise object localization but also generates bounding boxes that more accurately reflect the true dimensions of underwater litter, reducing false positives and increasing detection precision (Maharjan et al., 2022). By incorporating geometric attributes like aspect ratios and specific dimensions, EIoU offers robust performance, especially for diverse underwater litter shapes. Compared to CIoU, EIoU facilitates more effective gradient descent during training, accelerating convergence in complex visual tasks and enhancing real-time monitoring and generalization capabilities in underwater litter detection (Liu et al., 2024).

By incorporating these three innovative modules into YOLOv8, this research seeks to overcome the existing limitations of the model in detecting small objects in the natural environment. The RepVGG enhances the feature extraction layers, SimAM introduces an attention-based mechanism for highlighting essential features, and EIoU provides a refined measure for evaluating detection accuracy. Together, these improvements establish YOLOv8 as a more robust and versatile tool in the object detection arena, effectively addressing the intricate challenges associated with small object detection. The new architecture of the improved YOLOv8 network for underwater detection, modified from the classical YOLOv8, is shown in Fig. 4.

*2.5.2. Training parameters*

Detecting underwater litter in AASS images is challenging, as it involves both localization and classification tasks. For training models, the combined use of Binary Cross-Entropy (BCE) loss for classification and EIoU for bounding box regression is prevalent (Wang et al., 2023). The CE loss function is defined as formula (3).

$$L_{BCE} = -\frac{1}{N}\sum_{i=1}^{N}[y_i log(p_i) + (1-y_i)log(1-p_i)] \quad (3)$$

where $N$ is the total number of predictions, $y_i$ is the actual label, and $p_i$ is the predicted probability for the $i$-th prediction. This loss function efficiently handles the classification aspect of the detection, discerning underwater litter from the background by outputting confidence scores

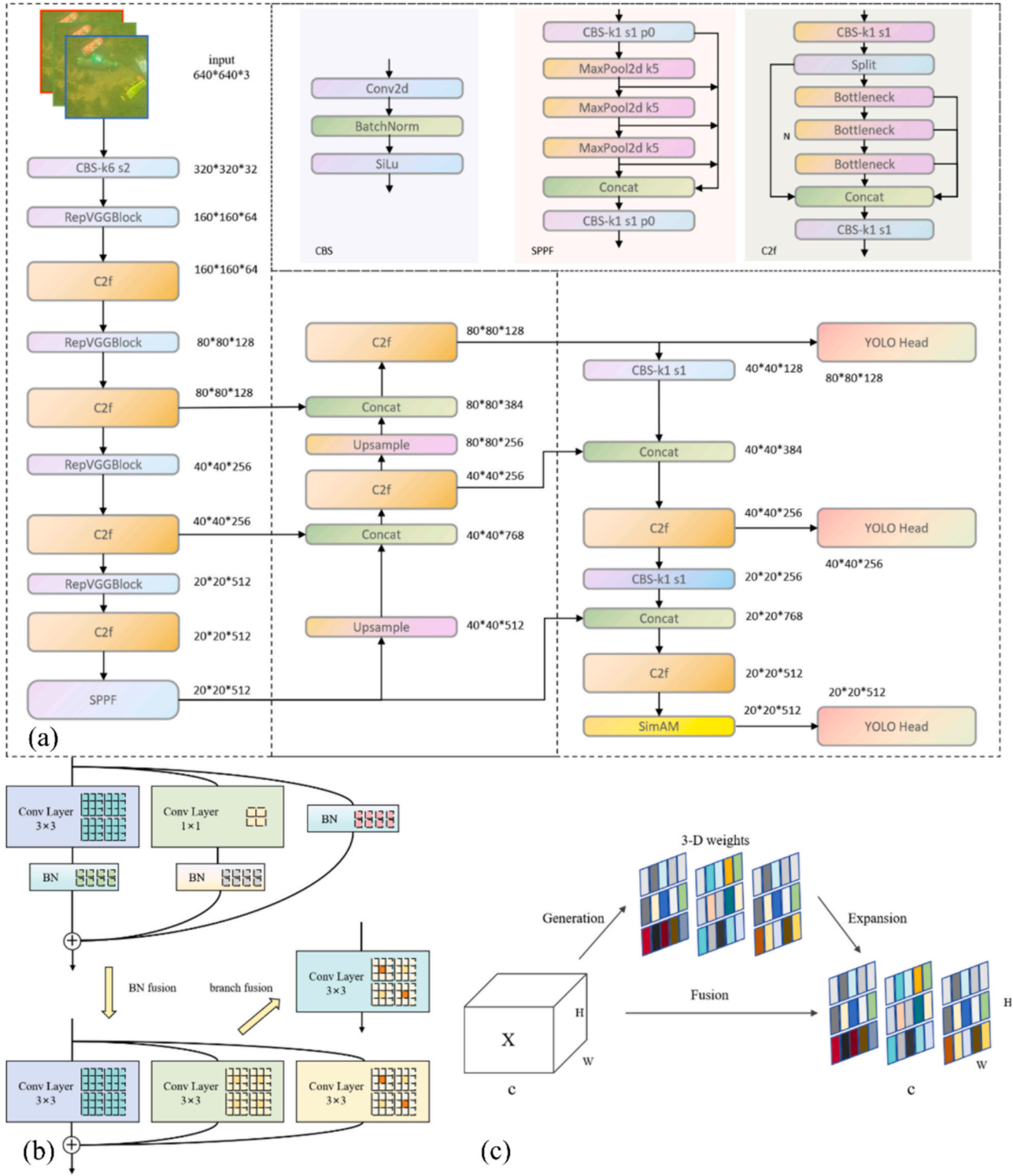

**Fig. 4.** Schematic diagram of the improved YOLOv8 architecture:(a) overall architecture; (b) RepVGG architecture; (c) SimAM architecture.

directly for each class and then taking the maximum score as the confidence for the anchor box.

For bounding box regression, RBL-YOLO employs the EIoU loss, which improves upon IoU by incorporating terms that account for the overlap area, the distance between the center points of the predicted and ground truth boxes, and the aspect ratio consistency. The EIoU loss function is defined as: EIoU

$$L_{EIoU} = 1 - IoU + \frac{\rho^2(b,\widehat{b})}{c^2} + \alpha v + \beta s \quad (4)$$

where $\rho^2(b,\widehat{b})$ calculates the euclidean distance between the predicted box $\widehat{b}$ and the ground truth box $b$, $c$ is the diagonal length of the smallest enclosing box covering both predicted and ground truth boxes, and $v$ measures the consistency of aspect ratio, with $\alpha$ being a positive trade-off parameter. RBL-YOLO uses a center-based method, predicting the distances from the center point to the left, top, right, and bottom edges of the bounding box, enhancing the model's ability to localize objects. Combining BCE and EIoU loss functions allow for a balanced approach to training the RBL-YOLO model for underwater litter detection in UAV images, addressing both classification and localization challenges. The total loss is computed as formula:

$$L_{Total} = \lambda_1 L_{BCE} + \lambda_2 L_{EIoU} \quad (5)$$

where $\lambda_1$ and $\lambda_2$ are weights that balance the contributions of the BCE and EIoU losses, respectively. This combined loss function, optimized via the Adam optimizer, significantly enhances the stability of the training process, ensuring the model effectively learns to identify and accurately localize underwater litter, despite their small size and complex backgrounds (Kingma and Ba, 2014).

### 2.5.3. Evaluation metrics

In the domain of object detection, evaluating the performance of models, such as those tasked with detecting underwater litter, involves a suite of metrics designed to quantify accuracy, reliability, and efficiency. A fundamental tool in this evaluation is the confusion matrix, which categorizes predictions into true positives (TP), true negatives (TN), false positives (FP), and false negatives (FN). These parameters form the basis for calculating key performance indicators. In this context, TP represents the count of pixels accurately identified as underwater litter, FP denotes the pixels incorrectly classified as underwater litter when they represent other objects, TN is the count of pixels correctly recognized as non-underwater litter entities, and FN refers to the pixels of actual underwater litter erroneously labeled as non-underwater litter objects. The F1-score, a harmonic mean of Precision and Recall, integrates both metrics to evaluate the model's comprehensive performance effectively.

$$Precision = \frac{TP}{TP+FP} \quad (6)$$

$$Recall = \frac{TP}{TP+FN} \quad (7)$$

$$F1-score = \frac{2 \times Precision \times Recall}{Precision + Recall} \quad (8)$$

$$mAP = \frac{1}{N}\sum_{i=1}^{N}\left(\sum_{j=0}^{n-1}(r_i - r_j)p_{interp}(r_{j+1})\right) \quad (9)$$

Within the scope of this study, N indicates the total number of unique object classes detected, which amounts to 2, reflecting the identification of underwater litter and non-litters. The variable n represents the number of recall levels used for calculating initial interpolated precision, arranged in ascending order. Symbols r and p stand for recall and precision, respectively. In comparison to the F1 score, the metric mean Average Precision (mAP) serves as a more robust evaluation mechanism for models tasked with multi-class detection. Unlike the F1 score, which offers a singular measure of performance, mAP provides an extensive assessment of a model's accuracy across different classes, rendering it a preferable metric for evaluating the effectiveness of object detection models in categorizing multiple object types.

## 3. Experimental results

### 3.1. Dataset description

Following the creation of a panoramic map of the riverbed, a detailed analysis was conducted using a sliding window technique, standardizing each extracted image to a dimension of 512 × 512 pixels. This process resulted in the segmentation of 1630 images from the collected field survey data. To enhance the model's learning efficiency, a series of data augmentation techniques was applied to the dataset. These techniques included rotation, symmetric flipping, and mirror reflection, all aimed at diversifying the dataset and bolstering the model's performance robustness.

In the pursuit of advancing object detection methodologies for underwater litter, this study employs an image degradation model to simulate LR conditions from HR imagery (Sert et al., 2019; Hayashi and Tsubouchi, 2022). The degradation model is mathematically represented as eq. (10):

$$g = (f \otimes h)\downarrow_s^{bicubic} + \eta \quad (10)$$

In this equation, $g$ symbolizes an LR image derived from the original HR image denoted by $f$. The term $h$ represents the point spread function, characterizing the blur introduced by uniform linear motion, while $\otimes$ signifies the convolution operation. The downsampling operation is indicated by $\downarrow$, with bicubic interpolation applied prior to downsampling by a factor of $s$, the magnification factor. $\eta$ accounts for the additive Gaussian white noise, introducing a level of randomness that mimics real-world imaging conditions.

The dataset underwent augmentation, increasing the number of images by a factor of 30, thus forming an enriched HR image set. Subsequently, the LR image dataset was generated through the application of the image degradation model to the HR dataset. Further diversification of the training set was achieved by extracting sub-images from the LR dataset. These sub-images, with dimensions of $l_{sub} \times l_{sub}$ pixels, were matched with corresponding HR sub-images cropped to $l_{sub} \times l_{sub}$, facilitating a comprehensive pairing of LR and HR samples for effective model training. After the processing, the datasets were divided into training, validation, and testing subsets.

### 3.2. Experimental framework

This study evaluates six SRR models—SRCNN, ESRGAN, EDSR, RDN, RCAN and SRFBN—on their ability to enhance the reconstruction quality of underwater litter images. Initially, these models were trained on a series of datasets and then applied to reconstruct SR images from an LR test set. The quality of these SR images was assessed using PSNR and SSIM. Additionally, a self-developed object detection network for this study was trained on HR underwater litter images and then used to detect the underwater litter using HR, bicubically upscaled (Bicubic), and SR images. The detection performance was evaluated using precision, recall, and mAP metrics.

### 3.3. SRR for underwater litter images

A series of networks were deployed to enhance the resolution of riverbed litter images through SRR. Table 1 displays comparisons of PSNR and SSIM for reconstructed images derived from seven methods. Results indicate that most of the deep learning-based SRR methods exhibit higher PSNR and SSIM compared to the conventional Bicubic method, suggesting superior image reconstruction. Among the selected networks, RCAN demonstrated superior performance, achieving the highest PSNR and SSIM values of 39.05 dB and 87.06 % respectively, indicative of its enhanced reconstructive capabilities. Notably, only the SRR images obtained by ESRGAN exhibit lower SSIM and PSNR values compared to the Bicubic images. The underlying reason might be attributed to the GAN-based nature of ESRGAN, likely due to their utilization of perceptual loss for better high-frequency detail reconstruction, in contrast to the L1 loss employed by CNN-based methods (Wang et al., 2018). Although CNN-based and GAN-based reconstructed images appear sharper than those generated by the Bicubic method, it is important to note that the high-frequency details in CNN-based images remain inadequate, rendering dense textures overly smooth (Wang et al., 2019). While GAN-based images seem more realistic, the generated textures still deviate from the reference images, which can be demonstrated in Fig. 5. The outlines of riverbed litter in SR image by RCAN were more distinctly delineated, and textures appeared significantly clearer, providing a more accurate representation of the underwater environment. However, it's worth noting that despite their advantages, deep learning-based methods do exhibit limitations, such as the potential for over-smoothing in certain textures, highlighting a trade-off between clarity and authenticity in reconstructed imagery. Nonetheless, the advancements these models bring to the task of SRR for riverbed litter images are undeniable, as they offer substantial improvements over traditional methods.

**Table 1**
Metrics of different SRR methods on the riverbed litter LR testset (x4).

| Metrics | Bicubic | ESRGAN | SRFBN | SRCNN | EDSR | RDN | RCAN |
| --- | --- | --- | --- | --- | --- | --- | --- |
| PSNR (dB) | 37.31 | 36.02 | 38.63 | 38.72 | 38.96 | 39.00 | **39.05** |
| SSIM (%) | 86.04 | 85.25 | 86.24 | 86.35 | 86.61 | 86.79 | **87.06** |

### 3.4. Object detection of reconstructed underwater litter images

Evaluation metrics for underwater detection were employed to assess the identification outcomes of both the Bicubic method and six deep learning-based SRR methods across all test sets. Table 2 presents the detection outcomes for the HR test set, Bicubic test set, and SR test sets generated by various algorithms. As depicted in Table 2, the HR test set attains the highest mAP of 79.6 %. The test set with the best reconstruction performance achieved by RCAN demonstrates the optimal performance among all SRR test sets, with a mAP of 78.6 %. The quality of the SRR images improves with the enhancement of super-resolution reconstruction metrics, leading to an increase in the comprehensive detection performance metric mAP as well as confidence score as shown in Fig. 6. In object detection, the confidence score is a metric that quantifies the certainty of the model regarding the presence of an object within a detected region. This score is a probability value ranging from 0 to 1, where a higher score indicates greater confidence that the detected object belongs to a specific class (Maji et al., 2022). Although the mAP of ESRGAN is 77.6 %, which is lower compared to other deep learning-based methods, it still surpasses the Bicubic method. This is because the SR images generated by ESRGAN possess more realistic textures, thereby enhancing detection performance compared to traditional Bicubic images.

Due to the diverse shapes, small sizes, and significant individual differences of underwater riverbed litter, coupled with the impact of image quality, this task is more challenging compared to traditional object recognition tasks. Therefore, this study conducted improvement experiments on the YOLOv8 model to adapt it for the target detection of underwater riverbed litter. The classic YOLOv8 model achieved a mAP of 0.76. Using this as a baseline, this study performed improvements on the backbone, attention mechanisms, and loss functions.

The classic YOLOv8 model uses CSPDarknet53 as the backbone, which is a variant of Darknet and uses Cross-Stage Partial (CSP) connections to enhance the flow of information between different stages of the network. In this study, RepVGG is used as a new backbone of RBL-YOLO. RepVGG is known for its streamlined convolutional network structure, which simplifies the architecture while maintaining or enhancing its representational capacity. RepVGG improves the model's ability to handle underwater debris datasets effectively, which are often characterized by common features in riverbed environments including uneven terrain, mixed with rocks, aquatic plants, and varying water quality. Additionally, RepVGG's ability to capture subtle feature variations enhances the model's accuracy in learning and categorizing various types of underwater debris. This makes RepVGG particularly suitable for the challenges of detecting underwater litter in riverbeds. As shown in Table 3, the recall, precision, and mAP are increased to 0.63, 0.77, and 0.79, respectively.

In the ablation experiment of the loss function, the classic YOLOv8 model generally uses CIoU. This study compares the performance of using Efficient Intersection over Union (EIoU), Distance Intersection over Union (DIoU), and Scale Intersection over Union (SIoU) as the YOLOv8 loss function. Among them, the model has the best effect when using EIoU as the loss function, obtaining a mAP of 0.80. Compared with the standard baseline of CIoU, the mAP of EIoU is increased by 3 %, and is improved by 5 % compared with SIoU. This is because EIoU considers the intersection ratio and bounding box alignment, thereby improving positioning accuracy; this is particularly important for underwater litter that varies greatly in size and shape (Thulasya Naik et al., 2024). The initial integration of SimAM caused a temporary drop in mAP due to changes in feature representation and the need for additional fine-tuning (Yang et al., 2021). However, by switching to a more effective loss function like EIoU, the model can achieve significant improvements in mAP. Fig. 7 shows the sample of distribution map of detected underwater litters, created by mosaicking a series of 512 × 512 pixel sub-images after prediction by the detection model. During the cropping process, some underwater litter samples are divided, resulting in detection failures. Additionally, a large plastic bag in the upper left corner appeared in multiple cropping sub-images due to its size, also leading to detection failure.

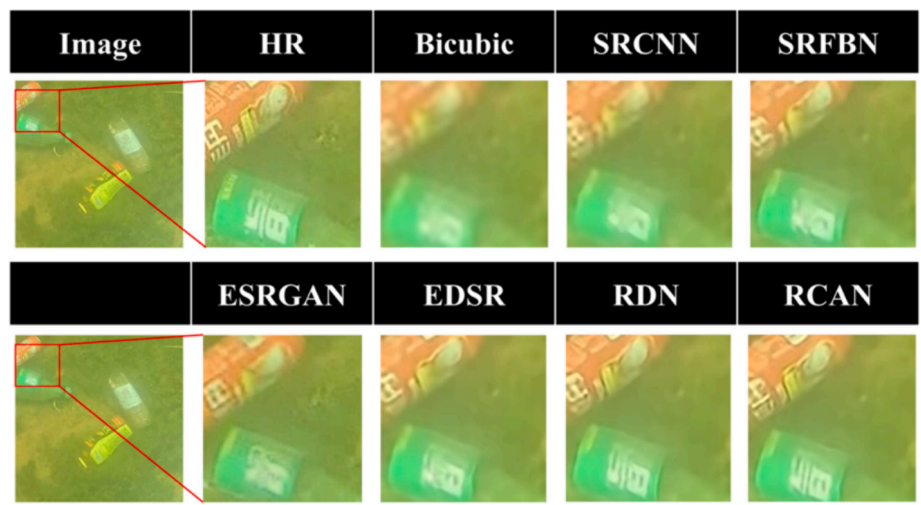

**Fig. 5.** Comparison of the visual effects of the reconstructed images of underwater litter using seven different methods.

**Table 2**
Evaluation metrics of detection results of the testset reconstructed by different SRR algorithms.

| Metrics (%) | HR | Bicubic | ESRGAN | SRFBN | SRCNN | EDSR | RDN | RCAN |
|---|---|---|---|---|---|---|---|---|
| mAP | 79.6 | 77.0 | 77.6 | 78.2 | 78.3 | 78.3 | 78.4 | 78.6 |

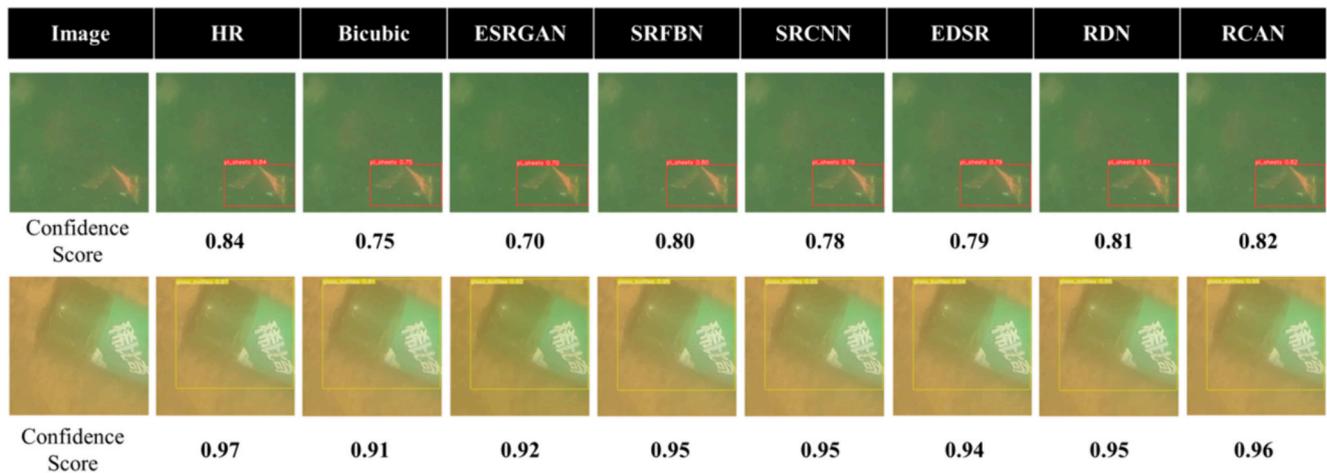

**Fig. 6.** Comparison of confidence scores of RBL-YOLO in target recognition on various SR test sets.

**Table 3**
Detection results for the different nets.

| Model | Precision | Recall | mAP |
|---|---|---|---|
| YOLOv8s | 0.89 | 0.57 | 0.76 |
| YOLOv8s-RepVGG | 0.63 | 0.77 | 0.79 |
| YOLOv8s-RepVGG-SimAM-EIoU | 0.86 | 0.70 | 0.80 |

A key challenge in underwater litter detection is identifying targets that are half-buried or hidden in sediment. To address this, sonar imaging is proposed, as it measures the reflectivity of sound waves to detect buried objects. Studies have shown its effectiveness in identifying items like mussel, which can similarly be applied to marine litter (Zhao et al., 2023a, b; Mizuno et al., 2022). For increased accuracy, manual verification through diver observations is recommended, particularly where sonar may be insufficient. Divers can compare findings with sonar data. In areas where manual validation isn't possible, AASS is suggested

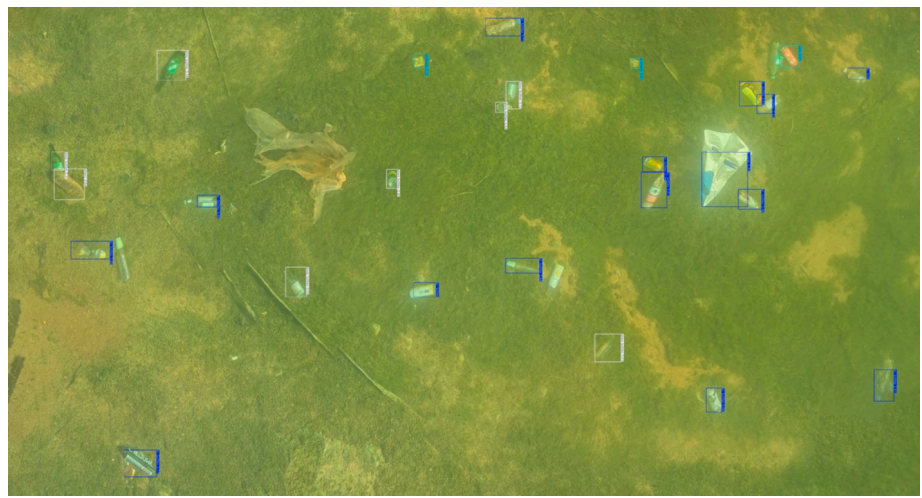

**Fig. 7.** Sample of distribution map of detected underwater litters.

to track objects over time due to its amphibious characteristics, improving detection reliability.

## 4. Discussion

### 4.1. The effect of magnification factor

In the application of SRR, the magnification factor is a critical parameter. It represents the resolution enhancement from LR images to SR images, reflecting the intensity of image enhancement achieved by SRR. To explore the influence of the SRR magnification factor and identify the optimal magnification factor for applying SRR technology in this task, the trained RCAN network was applied to the same LR dataset to generate the SR dataset at different magnification factors. The x1-LR test set (128 × 128) was then processed through the x2-SRR model to generate the x2-SR test set (256 × 256). Similarly, using corresponding magnification factors, the x3-SR (384 × 384), x4-SR (512 × 512), and x5-SR (640 × 640) test sets were generated. The study revealed that increasing the magnification factor generally enhances image resolution and detection accuracy, but also escalates computational demands. The recognition accuracy of the underwater litter images reconstructed by the SRR algorithm at different magnification factors using the RBL-YOLO model is shown in Table 4, and Fig. 8 shows the confidence scores of the RBL-YOLO model in target detection at different magnifications. Increasing the magnification factor can directly increase the resolution of SR images, which can usually improve the accuracy of advanced tasks such as object detection, but it will also lead to a significant increase in the computing resources required to train the model and output images (Xiang et al., 2022). In this experiment, as the magnification factor increases, the difference between the detection results of the SR images and the actual labels diminishes, though the demand for computing resources and time rises, the computation time for x4 and x5 increased by 20 % and 50 %, respectively, compared to x3. This shows the trade-off between improving detection accuracy and computational efficiency, as higher magnification factors lead to diminishing returns in accuracy improvements relative to the resource consumption. Adopting a magnification factor of 4 achieves the most favorable balance between detection accuracy and computational efficiency, as it offers the best trade-off between improvement in detection precision and the increase in computational time. This outcome is consistent with existing research on SRR across various magnification level (Xiang et al., 2022; Timofte et al., 2018; Li et al., 2023). This balance is critical in real-world applications where resource limitations are a concern, showcasing the importance of selecting an optimal magnification factor. After identifying the optimal magnification factor, the study demonstrated a clear improvement in detection accuracy at a factor of 4, providing a balance between precision and resource consumption. Large magnification factors and real-world conditions introduce the additional challenge of aligning high-resolution results with the ground truth at a subpixel level, which can lead to reduced accuracy, further highlighting the importance of selecting an optimal magnification factor (Timofte et al., 2018).

### 4.2. Effect of different object detection models on SRR

To examine the impact of various object detection networks on detecting underwater riverbed litter, 15 distinct models including two residual network-based models, and the YOLO series, such as s, n, spp., tiny, and CSP, were trained using a dataset specifically designed for high-resolution underwater litter detection. Analysis of the 16 networks' detection capabilities, as shown in Table 5, indicates that the model developed in this study surpasses other baseline models in detecting litter across all test sets. The YOLOv8 variations showcased enhanced mAP values across all datasets, proving the advantage of increased network depth. The newly proposed network outperforming the classical YOLOv8 by 4 % on the testset. This discrepancy is related to the model architecture; residual network-based models typically excel in image classification tasks but fall short in target detection tasks that require deep extraction of object details (Zoph et al., 2020; Li et al., 2022). Models with high precision but low recall are highly unfavorable for practical underwater litter monitoring because they disrupt the estimation of density distribution and affect the tracking of dynamic traceability of object (Chen et al., 2022; Liu et al., 2021). The proposed YOLO-RBL significantly improves the recall rate of the YOLO series models. Although the overall mAP metric has not shown substantial improvement, the model achieves a 13 % increase in recall compared to the YOLOv8s model, with only a 3 % decrease in precision. This suggests that the network's ability to generalize across different datasets contributes significantly to its improved recall performance.

### 4.3. Addressing data imbalance and object similarity in underwater litter detection

Fig. 9 shows the recall-precision curve of underwater litter object detection for each class. In the context of object detection, plastic bags (label: pl_bags) often appear as floating litter with variable shapes, resulting in a limited number of underwater samples. This study's dataset also exhibited an imbalance, with only two instances of plastic bags, leading to poor performance in detecting this category. Additionally, the visual similarity between glass bottles and plastic bottles contributes to misclassification errors. These issues of data imbalance and object similarity challenge the accuracy of current models calling for more diverse and representative training data.

To address these issues, the imbalance of plastic bag data can be mitigated by pre-training models on publicly available underwater litter datasets such as the TACO (Trash Annotations in Context), DeepPlastic and MARIDA (Marine Debris) dataset (Majchrowska et al., 2022; Kikaki et al., 2024). This approach can enhance the model's ability to recognize plastic bags by leveraging a larger and more diverse set of training examples. Another effective solution is to use CycleGAN (Almahairi et al., 2018; Sandfort et al., 2019) algorithm for data augmentation, involving water tank experiments to collect images of plastic and glass bottles in artificial environments. The CycleGAN can perform style transfer to generate synthetic images that mimic the appearance of underwater litter, increasing the quantity and variety of training data, thereby improving model performance (Jackson et al., 2019).

It is worth noting that some studies do not distinguish between plastic and glass bottles, classifying them together as "bottles" (Politikos et al., 2021; Sánchez-Ferrer et al., 2023; Majchrowska et al., 2022). However, this classification has limitations. Underwater litter mainly originates from household waste, such as beverage containers. Glass bottles, not being biologically toxic, can provide habitat for benthos such as *Barnacles* (Barnes et al., 1951). In contrast, plastic bottles can degrade into nano plastics, which accumulate in the food chain and pose significant ecological risks, including entangling small marine organisms (Bour et al., 2018). The implementation of these solutions can enhance the accuracy and reliability of underwater litter detection models, ensuring more precise classification and better management of marine debris. By addressing the data imbalance and object similarity, the proposed methods can contribute to the advancement of automated underwater litter detection technologies.

### 4.4. Prospects and scientific implications of AASS in future applications

The integration of the AASS into underwater litter detection marks an advancement in environmental monitoring. By combining consumer-grade aerial-aquatic drone technology with advanced imaging capabilities, the AASS addresses many challenges inherent in traditional survey methods. The potential of the AASS lies in its ability to enhance underwater litter detection and expand to other ecological monitoring tasks. Future work will explore potential improvements through the incorporation of underwater targets detection and acoustic video cameras, such as Adaptive Resolution Imaging Sonar (ARIS), and broader applications and implications of this technology (Zhao et al., 2023a, b; Mizuno et al., 2016). Acoustic video camera technology would complement the high-resolution acoustic imaging of the AASS, enhancing the system's adaptability and target detection capabilities in complex underwater environments.

**Table 4**
Evaluation metrics of detection results of the testset reconstructed by different magnification factors.

| Metrics (%) | HR | x1-LR | x2-SR | x3-SR | x4-SR | x5-SR |
| --- | --- | --- | --- | --- | --- | --- |
| mAP | 79.6 | 68.7 | 75.2 | 76.4 | **78.6** | 77.5 |

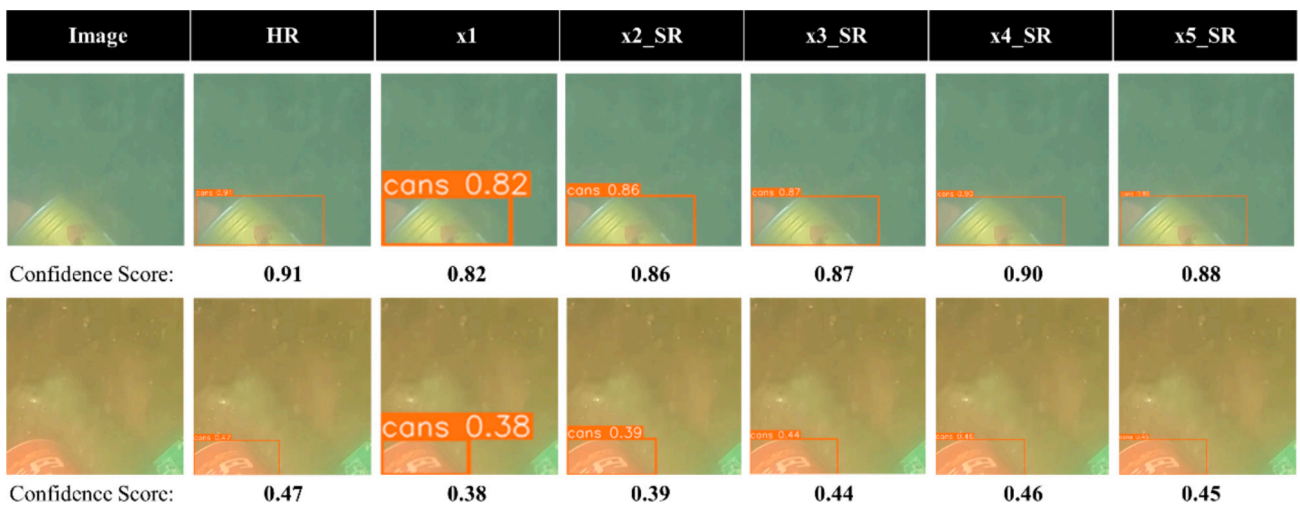

Fig. 8. Comparison of confidence score of object detection using RBL-YOLO on SR test sets of different magnifications factors.

**Table 5**
Underwater litter detection accuracy of different networks on the testset.

| Networks | Precision | Recall | mAP |
|---|---|---|---|
| ResNet50-csp | 0.35 | 0.21 | 0.18 |
| ResNeXt50-csp | 0.34 | 0.54 | 0.22 |
| YOLOv3-spp | 0.65 | 0.46 | 0.58 |
| v3 | 0.66 | 0.57 | 0.61 |
| v4-csp | 0.39 | 0.36 | 0.19 |
| v5s | 0.84 | 0.57 | 0.75 |
| v5n | 0.65 | 0.50 | 0.55 |
| v6s | 0.57 | 0.30 | 0.25 |
| v6n | 0.84 | 0.44 | 0.62 |
| YOLOr-csp | 0.28 | 0.44 | 0.15 |
| v7 | 0.86 | 0.55 | 0.63 |
| v7-tiny | 0.38 | 0.52 | 0.35 |
| v8n | 0.85 | 0.61 | 0.74 |
| v8s | **0.89** | 0.57 | 0.76 |
| **Ours** | 0.86 | **0.70** | **0.80** |

Integrating the AASS with instance segmentation algorithms in future marine litter recovery and management efforts could provide precise quantification of litter size and dimensions. This combination would offer valuable insights for underwater litter collection and dynamic tracing. The AASS with 4-meter beam and multi-camera setup can achieve a survey line width of approximately 15 m at a depth of around 5 m. The AASS is adaptable to depths ranging from 0.5 to 10 m. Coupled with a cruising speed of 1–2 knots and precise GPS information, the AASS holds significant advantages for marine litter monitoring in coastal areas. It offers detailed spatial mapping of underwater litter density and distribution with GPS information, surpassing previous underwater litter studies that focused solely on object detection accuracy without providing comprehensive distribution maps or industrial applications (Politikos et al., 2021; Fakiris et al., 2022).

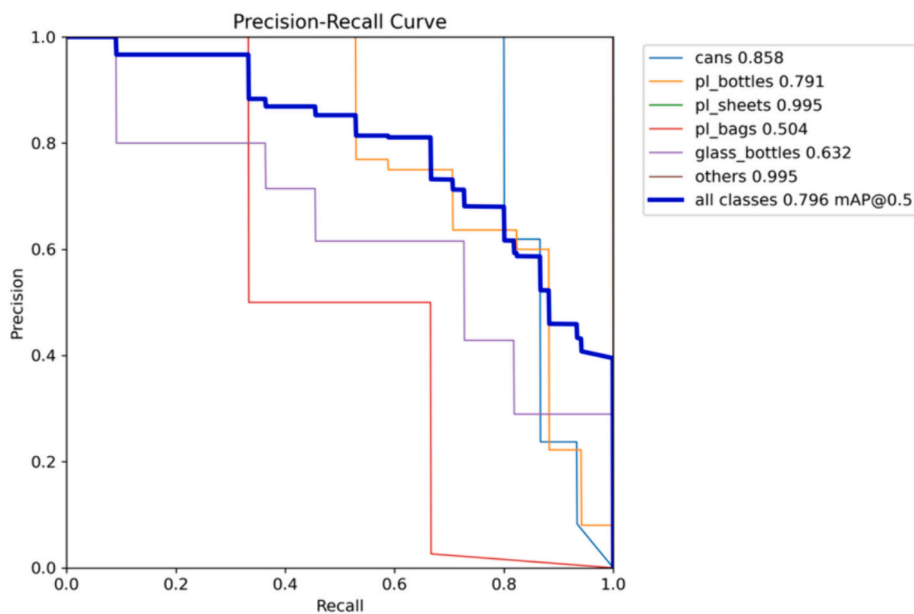

Fig. 9. Recall-precision curve of underwater litter object detection for each class.

The AASS has potential for broader ecological monitoring applications. Beyond litter detection, it could be used for coral cover assessment in reef flat zones or shallow waters, as well as for monitoring benthic species sensitive to environmental changes, such as sea cucumbers. Its aerial mode provides an overview of benthic habitats, facilitating the monitoring of marine life distribution and health. While numerous amphibious UAVs (Li et al., 2022; Rockenbauer et al., 2021; Space Entertainment Laboratory Co., Ltd, n.d.) and ROVs exist for underwater surveys, their high costs limit accessibility for small island nations and non-profit organizations crucial to environmental research. The AASS, with its consumer-grade components, offers a cost-effective solution, reducing scientific research inequality caused by equipment costs and making underwater target surveys more accessible and widely recognized.

Nevertheless, the AASS faces the challenge of turbidity, which can significantly reduce underwater visibility and affect data accuracy. To address this limitation, the use of sonar or ARIS technologies are promising solutions, which are less impacted by water quality and can effectively collect data in turbid waters. Additionally, algorithmic improvements, such as image deblurring and deturbidity methods like style transfer, are suggested to enhance image quality and mitigate the effects of turbidity.

## 5. Conclusion

This research proposes an automatic detection system for identifying such litter, utilizing deep learning-based SRR and object detection techniques. In terms of data acquisition, the AASS was developed, combining the efficiency of UAVs with the high-resolution imaging capabilities of AUVs. In terms of data processing, deep learning-based SRR was employed to refine LR images, enhancing their resolution and quality. This was followed by the application of the proposed RBL-YOLO object detection network, specifically designed for the accurate identification of underwater litter.

Underwater litter feature measurements within detection maps indicated that deep learning-based SRR techniques produce higher levels of image enhancement than the bicubic method. The precision in detecting and quantifying underwater litter features in SR images, refined using deep learning-based SRR, markedly exceeded the outcomes associated with images refined through the Bicubic method. The proposed RBL-YOLO model showcased superior accuracy and mAP on the HR dataset, outshining competing detection models. Given the broader environmental concerns, this system has the potential to improve waste management efforts in various aquatic environments, including rivers, lakes, and oceans. Moreover, the technology could be applied in large-scale environmental monitoring, aiding in global efforts to mitigate aquatic pollution and protect marine ecosystems.

Nevertheless, the two-stage approach for detecting underwater riverbed litter may not meet real-time requirements for super-resolution and detection. Future research will explore enhanced network designs that integrate SRR with object detection capabilities. Given the more complex conditions of coastal marine litter, upcoming studies will aim to advance and apply this technology to ocean underwater litter, enhancing the robustness and practicality of underwater litter surveys in challenging aquatic environments.

## CRediT authorship contribution statement

**Fan Zhao:** Writing – review & editing, Writing – original draft, Visualization, Validation, Supervision, Software, Resources, Project administration, Methodology, Investigation, Funding acquisition, Formal analysis, Data curation, Conceptualization. **Yongying Liu:** Writing – original draft, Methodology. **Jiaqi Wang:** Visualization, Software, Methodology, Investigation, Data curation. **Yijia Chen:** Resources, Project administration, Data curation. **Dianhan Xi:** Validation, Methodology. **Xinlei Shao:** Writing – review & editing, Validation, Methodology. **Shigeru Tabeta:** Supervision, Resources. **Katsunori Mizuno:** Supervision, Resources, Funding acquisition.

## Declaration of competing interest

The authors declare that they have no known competing financial interests or personal relationships that could have appeared to influence the work reported in this paper.

## Data availability

Data will be made available on request.

## Acknowledgements

We would like to express our gratitude to Miss. Yulun Chen (Department of Environmental Science, Southwest Forestry University) for her assistance on scientific research activities in Yuxi city, China. This work was partially supported by JST SPRING, Grant Number JPMJSP2108 and Windy Network corporation cooperative research funding.